\documentclass{bmvc2k}

\title{LSD$_2$ -- Joint Denoising and Deblurring of Short and Long Exposure Images with CNNs}

\addauthor{Janne Mustaniemi}{janne.mustaniemi@oulu.fi}{1}
\addauthor{Juho Kannala}{juho.kannala@aalto.fi}{2}
\addauthor{Jiri Matas}{matas@cmp.felk.cvut.cz}{3}
\addauthor{Simo S\"arkk\"a}{simo.sarkka@aalto.fi}{4}
\addauthor{Janne Heikkil\"a}{janne.heikkila@oulu.fi}{1}

\addinstitution{
 Center for Machine Vision and \\Signal Analysis, University of Oulu,\\
 Finland
}
\addinstitution{
 Department of Computer Science,\\
 Aalto University, Finland
}
\addinstitution{
 Center for Machine Perception,\\
 Faculty of Electrical Engineering,\\Czech Technical University in Prague,\\
 Czech Republic
}
\addinstitution{
 Department of Electrical \\Engineering and Automation,\\
 Aalto University,
 Finland
}
\runninghead{Mustaniemi et al.}{LSD$_2$ -- Joint Denoising and Deblurring}

\def\eg{\emph{e.g}\bmvaOneDot}

\def\etal{\emph{et al}\bmvaOneDot}

\usepackage{times}
\usepackage{epsfig}
\usepackage{graphicx}
\usepackage{amsmath}
\usepackage{amssymb}
\usepackage{booktabs}
\usepackage{dcolumn}
\usepackage[percent]{overpic}
\usepackage{comment}

\begin{document}

\maketitle

\begin{abstract}
The paper addresses the problem of acquiring high-quality photographs with handheld smartphone cameras in low-light imaging conditions. We propose an approach based on capturing pairs of short and long exposure images in rapid succession and fusing them into a single high-quality photograph. Unlike existing methods, we take advantage of both images simultaneously and perform a joint denoising and deblurring using a convolutional neural network. A novel approach is introduced to generate realistic short-long exposure image pairs. 
The method produces good images in extremely challenging conditions and outperforms existing denoising and deblurring methods. It also enables exposure fusion in the presence of motion blur. 
\end{abstract}

\section{Introduction}
Capturing high-quality images in difficult acquisition conditions is a formidable challenge. Such conditions, which are not uncommon, include low lighting levels and dynamic scenes with significant motion. Besides the problems of noise and motion blur, the camera sensors have limited dynamic range. Details are typically lost either in dark shadows or bright highlights. These problems are most pronounced in smartphones, where the camera and optics need to be small, light-weight and cheap. 

\begin{figure}
\begin{center}
\includegraphics[width=1.0\linewidth]{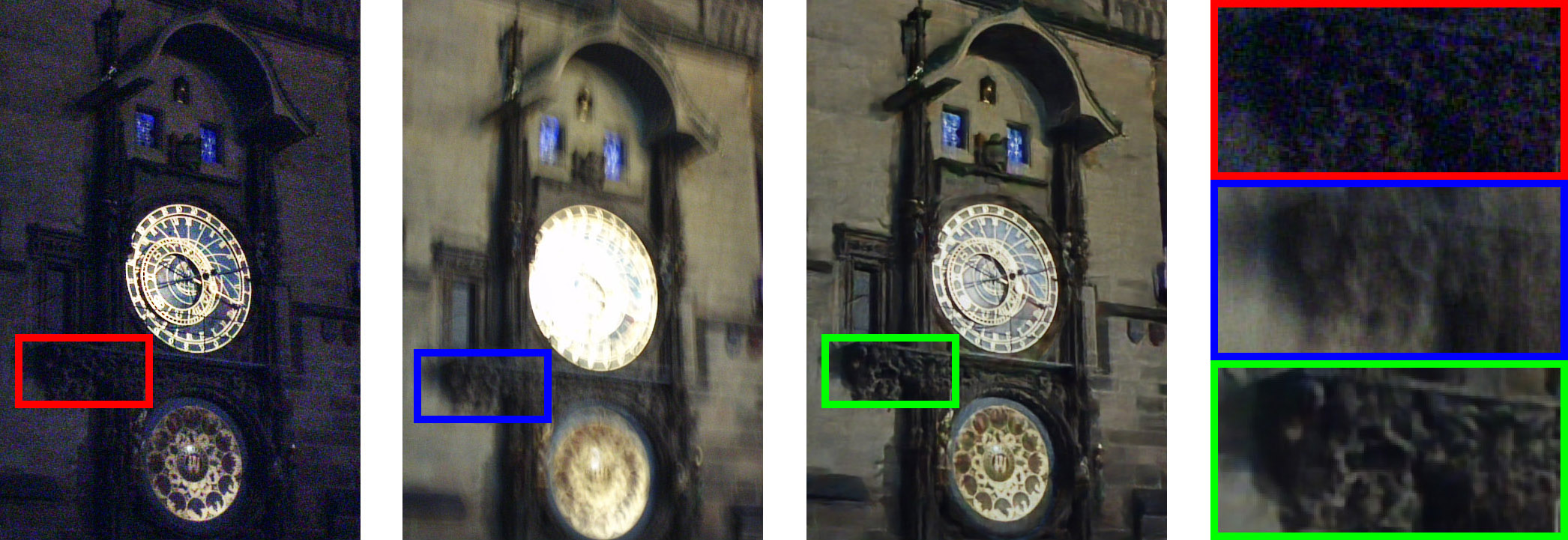}
\end{center}
\vspace{-2mm}
   \caption{A pair of short and long exposure images captured by a hand-held tablet at night in 7+210=217 milliseconds. A jointly deblurred and denoised image by the proposed LSD$_2$ method. The sharp and noise-free LSD$_2$ output has been tone mapped using an exposure fusion method. See Fig.~\ref{fig:results_dynamic} for dynamic scene results.}
  \label{fig:teaser}
\end{figure}

A satisfactory compromise between short and long exposure times does not always exist. To get rich colors, good brightness and low noise levels, one should choose long exposure with low sensor sensitivity setting. This will cause motion blur if the camera is moving (shaking), or if there is motion in the scene. A short exposure with high sensitivity setting will produce sharp but noisy images. Examples of such images are shown in Fig.~\ref{fig:teaser}. These issues can be addressed using image denoising and deblurring. However, the conventional methods are limited by the information in a single image.

We propose a novel approach that addresses the aforementioned challenges by taking ``the best of both worlds'' via computational photography. Current mobile devices can often be programmed to capture bursts of images without extra hardware or notable delay. We capture pairs of short and long exposure images and fuse them into a single high-quality image. The method does not rely on existing denoising algorithms unlike previous methods that utilize short-long exposure image pairs~\cite{yuan2007image,whyte2012non}.
  
The proposed method, called LSD$_2$\footnote{LSD$_2$ stands for Long-Short Denoising and Deblurring.} performs joint image denoising and deblurring, exploiting information from both images, adapting their contributions to the conditions at hand. We train a deep convolutional neural network (CNN) that takes a pair of short and long exposure images as input and provides a single image as output. A novel approach is proposed for generating realistic short- and long-exposure images. LSD$_2$ is shown to outperform existing single-image and multi-image methods. Additionally, we train a second network for exposure fusion. Processing the LSD$_2$ output with the exposure fusion network improves the colors and brightness compared to a single-exposure smartphone image. The LSD$_2$ network, training data, and the Android software we
developed for acquisition of the back-to-back short and long exposure images will be made public.

\section{Related work}
Single-image denoising has been addressed using approaches such as sparse representations \cite{elad2006image}, transform-domain collaborative filtering \cite{dabov2007image} or nuclear norm minimization \cite{gu2014weighted}. Several deep learning based approaches have been proposed recently \cite{jain2009natural,burger2012image,zhang2017beyond,lehtinen2018noise2noise}. The networks have been trained with pairs of clean and noisy images \cite{jain2009natural,burger2012image,zhang2017beyond}, and without clean targets \cite{lehtinen2018noise2noise}. Besides the end-to-end deep learning approaches, some utilize either conventional feed-forward networks \cite{zhang2017learning} or recurrent networks \cite{chen2017trainable} as learnable priors for denoising. 
Randomly initialized networks have been used as priors without pretraining \cite{deepimageprior}. Some methods utilize raw sensor data \cite{chen2018learning}. 
In contrast to our approach, the aforementioned methods focus on single image restoration and do not address multi-image denoising and deblurring. 


Single-image deblurring is an ill-posed problem. Various priors have been proposed to regularize the solutions, \eg the dark and bright channel priors \cite{pan2016blind,yan2017image}. These methods assume spatially invariant blur which limits their practicality. Priors based on deep networks have also been proposed \cite{zhang2017learning}. Some methods first estimate blur kernels and thereafter perform non-blind deconvolution \cite{sun2015learning,gong2017motion}. There are end-to-end approaches that directly produce a deblurred image \cite{nimisha2017blur,nah2017deep,DeblurGAN}. Some methods aim to remove deconvolution artifacts \cite{son2017fast,wang2018training}. Others utilize additional information like inertial measurements \cite{mustaniemi19,hee2014gyro}. Despite recent progress, single-image deblurring methods often fail to produce satisfactory results. Unlike our approach, they cannot utilize a sharp but noisy image to guide the deblurring.


Several multi-image denoising \cite{hasinoff2016burst,mildenhall2018burst} and deblurring approaches \cite{delbracio2015removing,wieschollek2016end,wieschollek2017learning,aittala2018burst} have been proposed recently. They process a burst of images captured with a constant exposure time. Therefore, they address either denoising or deblurring, but not both problems jointly like we do. As the input images are not as complementary, they cannot get ``the best of both worlds'' but suffer the drawbacks of either case. 
Dynamic scenes are difficult to handle, especially when the capture time is long. Images may be severely misaligned and fast-moving objects might disappear from the view. With a constant exposure, the saturated regions can not be easily avoided and high dynamic range imaging is not achieved. On top of that, based on our observations and earlier studies \cite{mildenhall2018burst,aittala2018burst}, it seems that due to the non-complementary nature of constant exposure, it is necessary to use more input frames than two. This may increase the consumption of memory, power, and processing time. 

A similar problem setting as in our work is considered in \cite{yuan2007image,whyte2012non}. These methods utilize short-long exposure image pairs for image deblurring. They first estimate blur kernels for the blurry image and thereafter use the so-called residual deconvolution, proposed by \cite{yuan2007image}, to iteratively estimate the residual image that is to be added to the denoised sharp image. A non-uniform blur model was introduced in \cite{whyte2012non} to improve the results. We note that both methods use \cite{portilla2003} for denoising. A significant drawback of \cite{yuan2007image} and \cite{whyte2012non} is that they require a separate photometric and geometric registration stage, where the rotation is estimated manually \cite{yuan2007image}. Moreover, their model is not applicable to non-static scenes. 

\section{Joint Denoising and Deblurring}
\begin{figure}
\begin{center}
\includegraphics[width=1.0\linewidth]{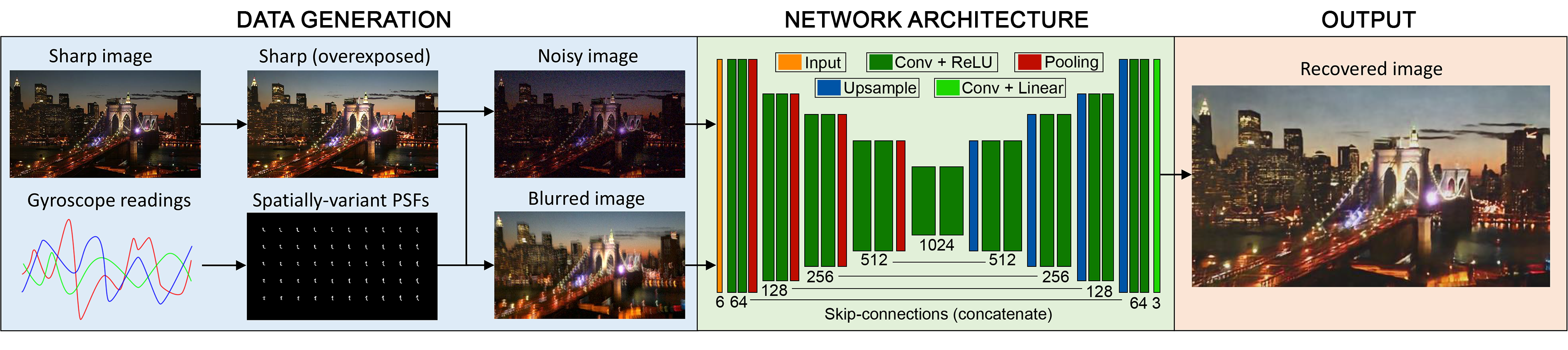}
\vspace{-8mm}
\end{center}
   \caption{Overview of the LSD$_2$ joint denoising and deblurring method. To train the network, we generate pairs of short and long exposure images with realistic motion blur, spatial misalignment, image noise, color distortion and saturated regions.}
\label{fig:overview}
\end{figure}

An overview of the proposed LSD$_2$ method is shown in Fig.~\ref{fig:overview}. The goal is to recover the underlying sharp and noise-free image using a pair of long and short exposure images. A short exposure image is sharp but noisy as it is taken with a high sensitivity setting. A long exposure image is typically blurry due to camera or scene motion. Note that colors of the short exposure image are often distorted w.r.t. the long exposure image as shown in Fig. \ref{fig:teaser}. Furthermore, the images are slightly misaligned even though they are captured immediately one after the other.




\subsection{Data Generation}
\label{sec:data_generation}
In order to train the network, we need pairs of noisy and blurry images together with the corresponding sharp images. Since there is no easy way to capture such real-world data, we propose a data generation framework that synthesizes realistic pairs of short and long exposure images. By utilizing images taken from the Internet and gyroscope readings, we can generate unlimited amount of training data with realistic blur while covering a wide range of different scene types. The LSD$_2$ network operates with images having intensity range $[0,1]$ and hence we first scale the original RGB values to that range. We first invert the gamma correction of the input images (assuming $\gamma = 2.2$). Once the images have been generated, the gamma is re-applied.

\subsubsection{Synthesis of Long Exposure Images}
\label{sec:saturation}
We take a regular high-quality RGB image $I$ from Internet as the starting point of our simulation. At test time, the long exposure image should be slightly overexposed in order to enable high dynamic range and ensure sufficient illumination of darkest scene regions. Hence, we simulate the saturation of intensities due to overexposure. We do that by first multiplying the intensity values with a random number $s$ uniformly sampled from the interval $[1,3]$. Then, by clipping the maximum intensity to value of 1, we get the sharp long exposure image, which will be the ground truth target. That is, we train the network to predict an output with similar exposure as the long exposure image.

\subsubsection{Underexposure and Color Distortion}
\label{sec:color_distortion}
The underexposed short exposure image is synthesized from the aforementioned long exposure image $sI$, where intensities can exceed 1, by applying affine intensity change ($asI+b$) with random coefficients ($a,b$) sampled from uniform distributions, whose parameters are determined by analyzing the intensity distributions of real short and long exposure pairs. The real image pairs are captured so that the short exposure time is always 1/30 of the long exposure time. Based on our observations, a constant exposure time ratio improves the performance. The absolute exposure time is allowed to vary based on auto-exposure algorithm.

The colors are often distorted in the noisy short exposure image as show in Fig.~\ref{fig:teaser}. Hence, in order to simulate the distortion, we randomly sample different affine transformation parameters ($a_i,b_i$) for each color channel $i$. Moreover, the parameters of the uniform distributions for $a_i$ and $b_i$ are determined independently for each color channel and they are such that $a_i<0.3$ and $b_i<0.01$ always. By introducing random color distortions, we encourage the network to learn the colors and brightness mainly from the (blurry) long exposure image. The final short exposure image is obtained by adding noise as described in Sec.~\ref{sec:noise}.

\subsubsection{Motion Blur}
\label{sec:motion_blur}
The motion blur is simulated only to the long exposure image $sI$.
Synthetically blurred images are generated with help of gyroscope measurements. Similar to prior work \cite{hee2014gyro,vsindelavr2013image}, we assume that motion blur is mainly caused by the rotation of the camera. We start by recording a long sequence of gyroscope readings with a mobile device. The idea is to simulate a real life imaging situation with a shaking hand.

The starting time of the synthetic image exposure $t_1$ is randomly selected to make each of the blur fields different. The level of motion blur is controlled by the exposure time parameter $t_e$, which defines the end time of the exposure $t_2 = t_1 + t_e$. The rotation of the camera $\mathbf{R}(t)$ is obtained by solving the quaternion differential equation driven by the angular velocities and computing the corresponding direction cosine matrices \cite{Titterton+Weston:2004}.
Assuming that the translation is zero (or that the scene is far away), the motion blur can be modelled using a planar homography 
\begin{equation}
\mathbf{H}(t) = \mathbf{K} \mathbf{R}(t) \mathbf{K}^{-1},
\label{eq:homography_simple}
\end{equation}
where $\mathbf{K}$ is the intrinsic camera matrix. Let $\mathbf{x}=(x,y,1)^\top$ be a projection of the 3D point in homogeneous coordinates. The point-spread-function (PSF) of the blur at the given location can be computed by $\mathbf{x}' = \mathbf{H}(t) \mathbf{x}$.

Since mobile devices are commonly equipped with a rolling shutter camera, each row of pixels is exposed at slightly different time. This is another cause of spatially-variant blur \cite{su2015rolling}. When computing the PSFs, the start time of the exposure is adjusted based on the y-coordinate of the point $\mathbf{x}$. Let $t_r$ denote the camera readout time, i.e. the time difference between the first and last row exposure. The exposure of the $y$:th row starts at $t_1(y) = t_f + t_r \frac{y}{N}$, where $t_f$ corresponds to the starting time of the first row exposure and $N$ is the number of pixel rows. To take this into account, we modify Eq.~\ref{eq:homography_simple} so that
\begin{equation}
\mathbf{H}(t) = \mathbf{K} \mathbf{R}(t) \mathbf{R}^{\top}(t_1) \mathbf{K}^{-1}.
\label{eq:homography_rolling}
\end{equation}
The blurred image is produced by performing a spatially-variant convolution between the sharp image and PSFs.

\subsubsection{Spatial Misalignment}
\label{sec:spatial_misalignment}
It is assumed that the blurry image is captured right after the noisy image. Nevertheless, the blurry image might be misaligned with respect to the noisy image due to camera or scene motion. 
Normally, the origin of the PSF would be at the center of the kernel (middle of the exposure). To introduce the effect of spatial misalignment, we set the origin of each PSF kernel to be at the beginning of the exposure. 
This approach also extends to cases when there is a known gap between the two exposures.

\subsubsection{Realistic Noise}
\label{sec:noise}
As a final step, we add shot noise to both generated images. The shot noise is considered to be the dominant source of noise in photographs, modeled by a Poisson process. The noise magnitude varies across different images since it depends on the sensitivity setting (ISO) of the camera. In general, the noise will be significantly more apparent in the short exposure image, and we model this by setting the noise magnitude for the short exposure image larger by a constant factor of 4. After this, we ensure that the maximum intensity of the blurry long exposure image does not exceed the maximum brightness value of 1. That is, we clip larger values at 1. Later in Sec.~\ref{sec:fine_tuning}, the network is fine-tuned with real examples of noisy images. This way the noise characteristics can be learned directly from the data. 

\subsection{Network and Training Details}
\label{sec:architecture}
The network is based on the popular U-Net architecture \cite{ronneberger2015u}. This type of network has been successfully used in many image-to-image translation problems \cite{pix2pix2017}. It was chosen because of its simplicity and because it produced excellent results for this problem. The architecture of the network is shown in Fig.~\ref{fig:overview}. The input of the network is a pair of blurry and noisy images (stacked). The images can be of arbitrary size since the network is fully convolutional. 
The number of feature maps is shown below the layers. All convolutional layers use a 3x3 window, except the last layer, which is a 1x1 convolution. Downsampling layers are 2x2 max-pooling operations with a stride of 2.

The LSD$_2$ network was trained on 100k images taken from an online image collection \cite{huiskes2010new}. The synthetically corrupted images have resolution of 270 $\times$ 480 pixels. We used the Adam \cite{kingma2015j} optimizer with the L2 loss function. The learning rate was initially set to 0.00005 and it was halved after every 10th epoch. The network was trained for 50 epochs.

\subsubsection{Fine-tuning}
\label{sec:fine_tuning}
The method is targeted for images that have gone through an unknown image processing pipeline of the camera. To this end, we fine-tune the network with real images captured with the NVIDIA Shield tablet. This way, the network can learn the noise and color distortion models directly from the data. Real noise has a relatively coarse appearance as can be seen in Fig.~\ref{fig:results_static}.
Our synthetic noise model assumes that the noise is independent for each pixel. This clearly does not hold because of the camera's internal processing (demosaicing, etc.).

We capture pairs of short and long exposure images while the camera is on a tripod. The ratio of exposure times is fixed to 1/30. The ISO settings for the long and short exposure images are set to 200 and 800, respectively. The long exposure image is used as the ground truth sharp image and the short exposure image directly corresponds to the noisy image. The blurred image is generated from the sharp image as described in Sec. \ref{sec:motion_blur}. 

To increase the amount of training samples, we capture several image pairs at once while varying the long exposure between 30 - 330 milliseconds. Moreover, the original images are divided to four sub-images of size 480 x 960 pixels. The network was fine-tuned on 3500 images for 30 epochs. The rest of the details are the same as in the previous section.

\section{Experiments}
We capture pairs of noisy and blurry images in rapid succession with the NVIDIA Shield tablet and the Google Pixel 3 smartphone. The image acquisition setup is the same as in Sec. \ref{sec:fine_tuning}, except this time the camera and/or scene is moving. The resolution of the images is 800 $\times$ 800 pixels (cropped from the original images). For the quantitative comparison, we use synthetically blurred and noisy image pairs taken from the validation set. An example of such pair is shown in Fig. \ref{fig:overview}.

\subsection{Single-Image Approaches}
\begin{figure}[!hbt]
  \centering
  \begin{overpic}[width=1.0\linewidth]{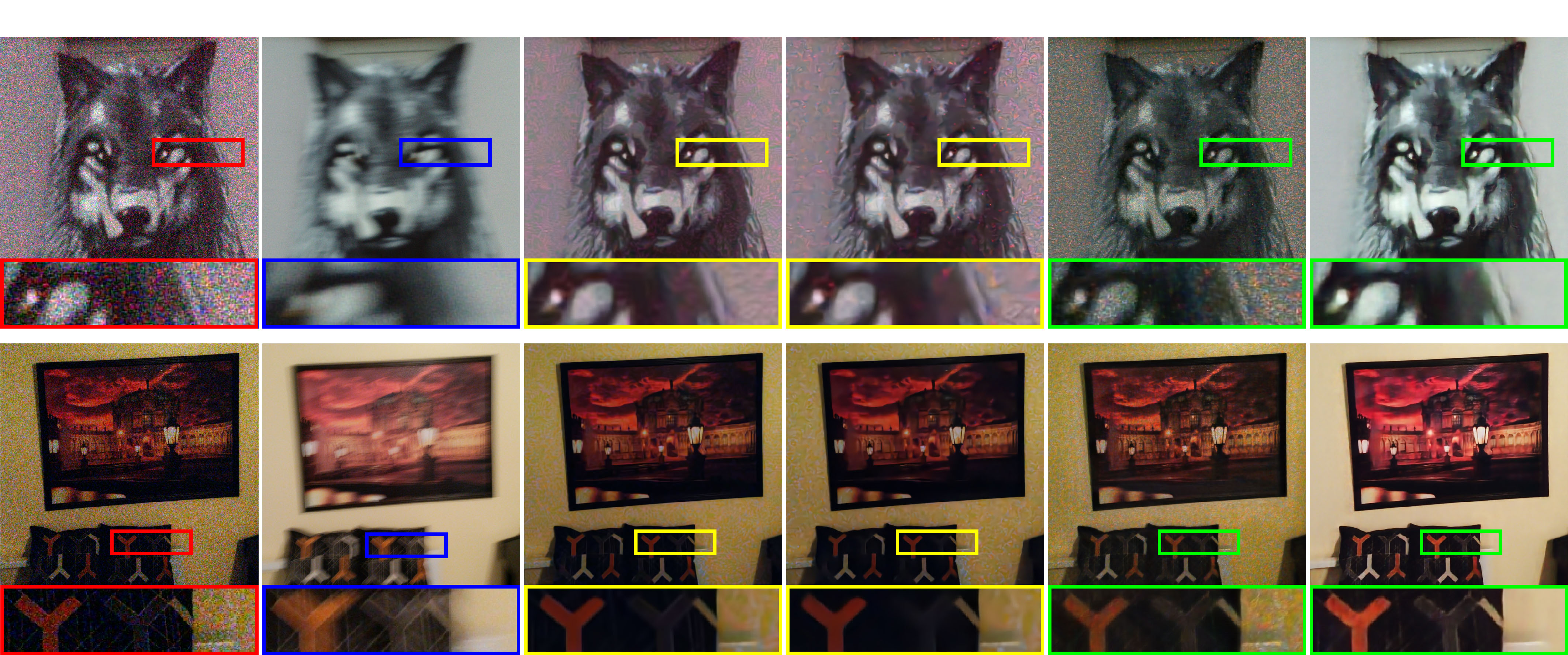}
  \put (6.0,40.1) {\small Noisy}
  \put (21.8,40.1) {\small Blurry}
  \put (36.7,40.1) {\small BM3D \cite{dabov2007image}}
  \put (51.5,40.1) {\small FDnCNN \cite{zhang2017beyond}}
  \put (68,40.1) {\small LSD$_2$ (synth.)}
  \put (89.5,40.1) {\small LSD$_2$}
  \end{overpic}
  \caption{A comparison of LSD$_2$ and single-image denoising methods BM3D \cite{dabov2007image} and FDnCNN \cite{zhang2017beyond}. The second column from the right shows the results without fine-tuning (synthetic data only). Note that the LSD$_2$ network was fine-tuned using NVIDIA Shield data but the input images on the second row were captured with Google Pixel 3. }
  \label{fig:results_static}
\end{figure}

\begin{figure}
  \vspace{1mm}
  \centering
  \begin{overpic}[width=1.0\linewidth]{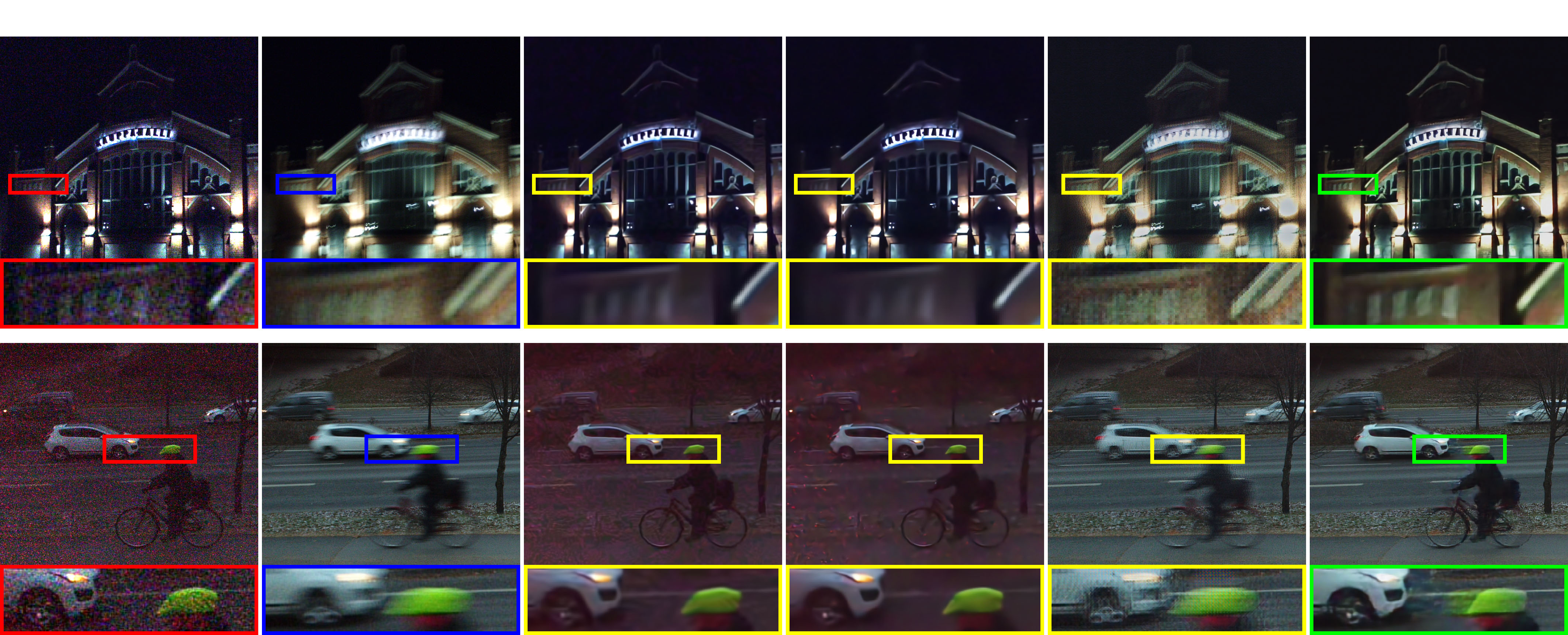}
\put (6.0,39.0) {\small Noisy}
  \put (22.0,39.0) {\small Blurry}
  \put (36.8,39.0) {\small BM3D \cite{dabov2007image}}
  \put (51.0,39.0) {\small FDnCNN \cite{zhang2017beyond}}
  \put (67.2,39.0) {\small DeblurGAN \cite{DeblurGAN}}
  \put (89,39.0) {\small LSD$_2$}
  \end{overpic}
  \caption{A low light performance in the presence of saturated pixels (top) and a dynamic scene performance (bottom).}
  \label{fig:results_dynamic}
\end{figure}
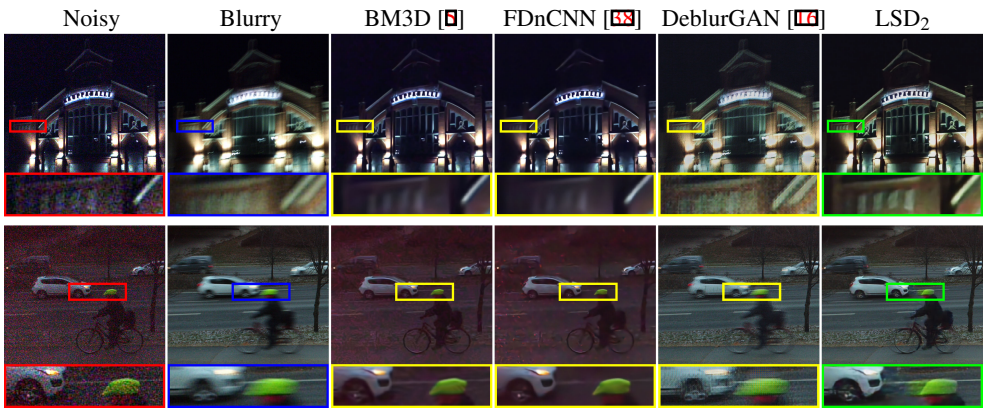

A comparison between LSD$_2$ and the state-of-the-art denoising methods BM3D \cite{dabov2007image} and FDnCNN \cite{zhang2017beyond} is shown in Figs \ref{fig:results_static} and \ref{fig:results_dynamic}. The short exposure image (noisy) has been normalized so that its intensity matches the blurry image for visualization. The most apparent weakness of BM3D and FDnCNN is that the color information is partly lost and cannot be recovered using a noisy image alone. Both methods tend to over-smooth some of the details even though their noise standard deviation parameters have been manually tuned to achieve a good overall balance between noise removal and detail preservation. A quantitative comparison in Table \ref{tab:results_quantitative} shows that LSD$_2$ outperforms the other methods by a fair margin.


Fig.~\ref{fig:results_dynamic} show a comparison against the state-of-the-art deblurring method \cite{DeblurGAN}. The results of DeblurGAN are unsatisfactory as it fails to remove most of the blur. Note that saturated image regions, such as light streaks, do not cause problems for LSD$_2$. Furthermore, LSD$_2$ performs surprisingly well on a dynamic scene even though it has not been trained for this type of situations. However, fine details such as the bike wheels remain blurry. See the supplementary material for more results.

We also tried retraining FDnCNN \cite{zhang2017beyond} and DeblurGAN \cite{DeblurGAN} using our dataset but with poor results. The FDnCNN network has to learn to adjust the colors and brightness in addition to denoising. This seems to be too challenging task at least for the given network. Retraining did not improve DeblurGAN likely because the blurred and sharp images are misaligned in our dataset.

The importance of using real data for fine-tuning is demonstrated in Fig.~\ref{fig:results_static}. The fine-tuning clearly helps as the output is significantly less noisy and the colors are better. Furthermore, the fine-tuning does not make the network device specific. 


\begin{table}
\begin{center}
\setlength{\tabcolsep}{6.5pt}
\begin{tabular}{|l|c|c|c|c|c|c|}
\hline
 & Noisy & Blurred & DeblurGAN \cite{DeblurGAN} & BM3D \cite{dabov2007image} & FDnCNN \cite{zhang2017beyond} & LSD$_2$ \\
\hline
PSNR & 16.43 & 16.88 & 15.78 & 23.48 & 23.83 & \bf 25.67 \\
SSIM & 0.51 & 0.57 & 0.54 & 0.79 & 0.81 & \bf 0.89 \\
\hline
\end{tabular}
\end{center}
\caption{The average peak-signal-to-noise ratio (PSNR) and structural similarity (SSIM) computed for 30 synthetically corrupted image pairs (shown in the supplementary material). For fairness, the outputs of \cite{dabov2007image} and \cite{zhang2017beyond} have been adjusted so that the colors match the blurred images before computing the scores.}
\label{tab:results_quantitative}
\end{table}

\subsection{Multi-Image Approches}

\begin{figure}
  \centering
  \begin{overpic}[width=1.0\linewidth]{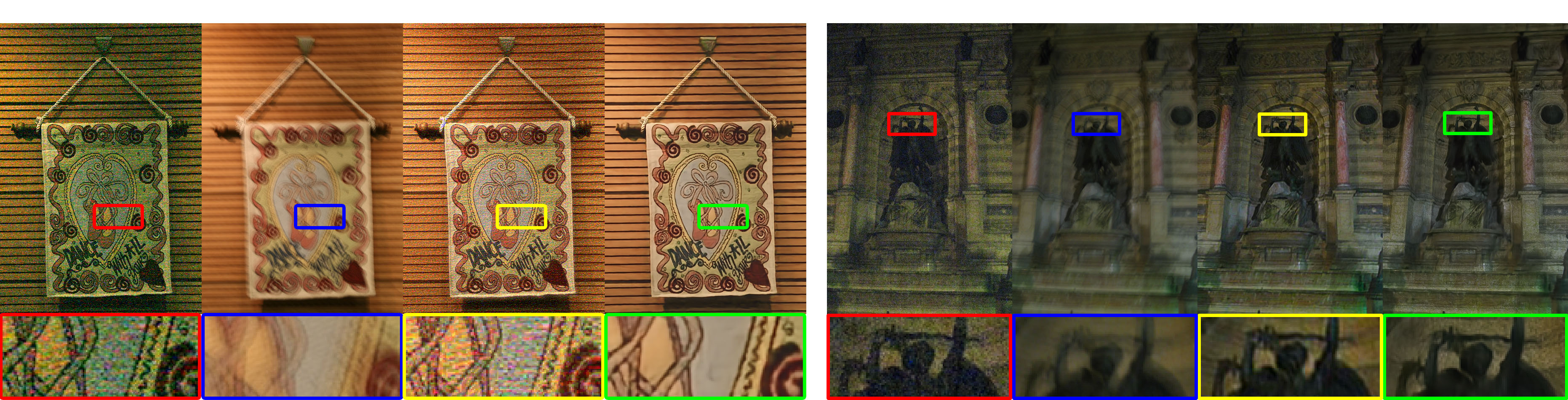}
  \put (4.0,24.7) {\small Noisy}
  \put (16.1,24.7) {\small Blurry}
  \put (30.5,24.7) {\small \cite{yuan2007image}}
  \put (42.7,24.7) {\small LSD$_2$}
  \put (56.3,24.7) {\small Noisy}
  \put (67.7,24.7) {\small Blurry}
  \put (80.0,24.7) {\small \cite{whyte2012non}}
  \put (91.7,24.7) {\small LSD$_2$}
  \end{overpic}
  \caption{A comparison with Yuan \etal \cite{yuan2007image} and Whyte \etal \cite{whyte2012non}. They both require a separate denoising step and manual alignment.}
  \label{fig:results_whyte}
\end{figure}
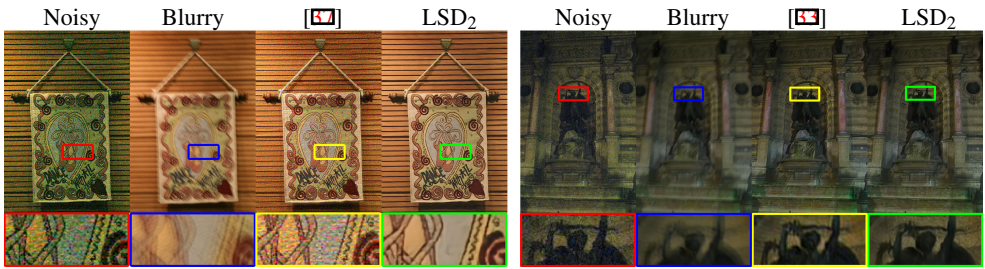

Fig.~\ref{fig:results_whyte} shows a comparison against Yuan \etal \cite{yuan2007image}. Input images were copied from the original paper (PDF file) as their inputs are not available. The LSD$_2$ output is significantly less noisy. The performance of \cite{yuan2007image} depends heavily on the separate denoising step. Furthermore, it produces major artifacts when inputs are misaligned (supplementary material). The parameters of \cite{yuan2007image} were kept default. Note that LSD$_2$ does not have any tunable parameters.

The implementation of Whyte \etal \cite{whyte2012non} is not publicly available. For comparison, we use the input images provided by the authors of \cite{whyte2012non}. Fig. \ref{fig:results_whyte} shows a comparison against the original result by \cite{whyte2012non}. LSD$_2$ produces equally good if not better results. The output of \cite{whyte2012non} shows a little bit of ringing and slightly less details. Keep in mind that \cite{whyte2012non} and \cite{yuan2007image} utilize an existing denoising algorithm. Moreover, the inputs need to be manually registered.

\begin{figure}
  \vspace{2mm}
  \centering
  \begin{overpic}[width=1.0\linewidth]{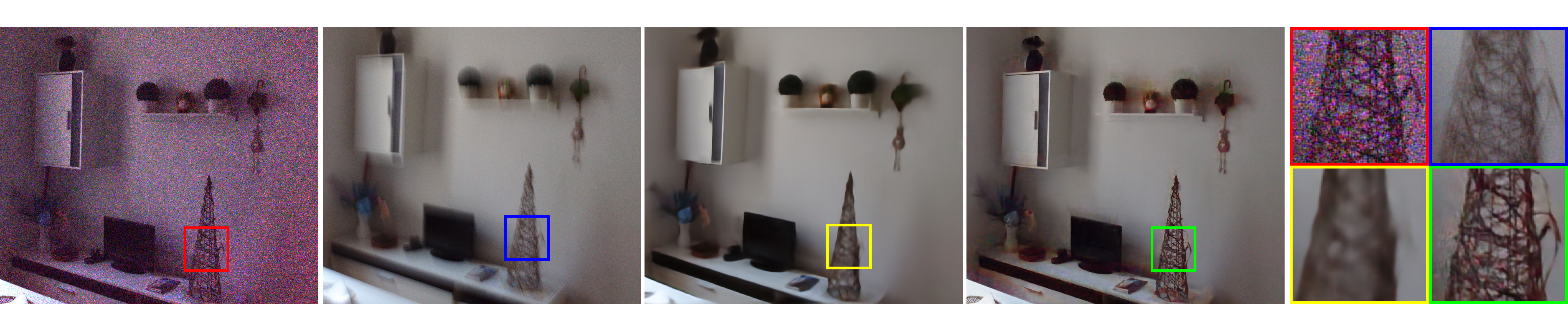}
  \put (7.5,18.9) {\small Noisy}
  \put (27.5,18.9) {\small Blurry}
  \put (40.5,18.9) {\small Aittala and Durand \cite{aittala2018burst}}
  \put (70.0,18.9) {\small LSD$_2$}
  \put (87.5,18.9) {\small Details}
  \end{overpic}
  \caption{A comparison with Aittala and Durand \cite{aittala2018burst}. A burst of 6 blurry images was given to \cite{aittala2018burst} as input (supplementary material).}
  \label{fig:results_aittala}
\end{figure}

A recent burst deblurring method by Aittala and Durand \cite{aittala2018burst} takes an arbitrary number of blurry images as input. Using their implementation, we compare the methods in Fig.~\ref{fig:results_aittala}. The final result appears less sharp compared to ours, which is obtained with only two images (blurry and noisy). Furthermore, the saturated regions such as the overexposed windows, cannot be recovered using the long exposure images alone. We also tried feeding a pair of noisy and blurry images to \cite{aittala2018burst} but the results were poor. This is not surprising as their method is designed for blurry images only. Similar to \cite{whyte2012non,yuan2007image}, the input images need to be registered in advance.
 
\begin{figure}
  \vspace{4mm}
  \centering
  \begin{overpic}[width=1.0\linewidth]{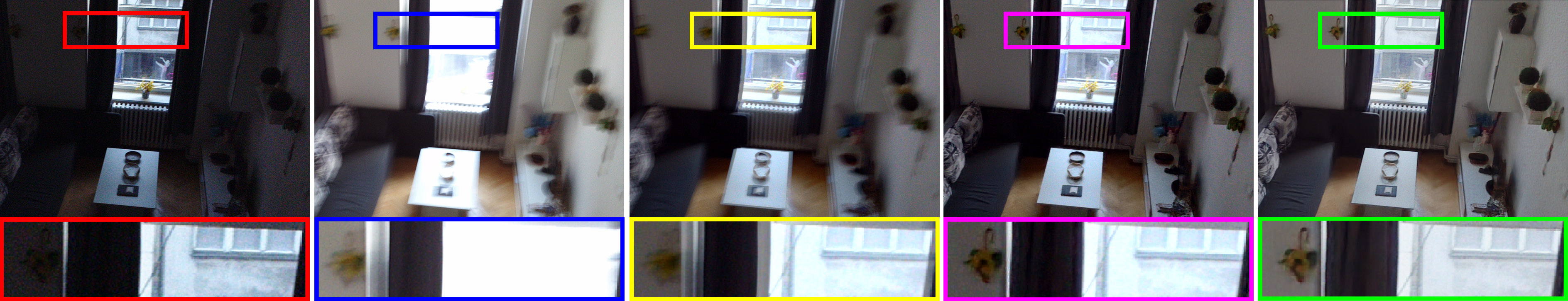}
  \put (7.5,20.3) {\small Noisy}
  \put (27.5,20.3) {\small Blurry}
  \put (43.0,20.3) {\small DeepFuse \cite{prabhakar2017deepfuse}}
  \put (63.5,20.3) {\small LSD$_2$ + \cite{prabhakar2017deepfuse}}
  \put (84.0,20.3) {\small LSD$_2$ + EF}
  \end{overpic}
  \caption{Exposure fusion  without and with LSD$_2$. DeepFuse \cite{prabhakar2017deepfuse}: using a pair of noisy and blurry images as input. LSD$_2$ + DeepFuse \cite{prabhakar2017deepfuse}: the blurry image is replaced with the LSD$_2$ output. LSD$_2$ + EF: using our exposure fusion method instead of \cite{prabhakar2017deepfuse}.}
  \label{fig:exposure_fusion}
\end{figure}
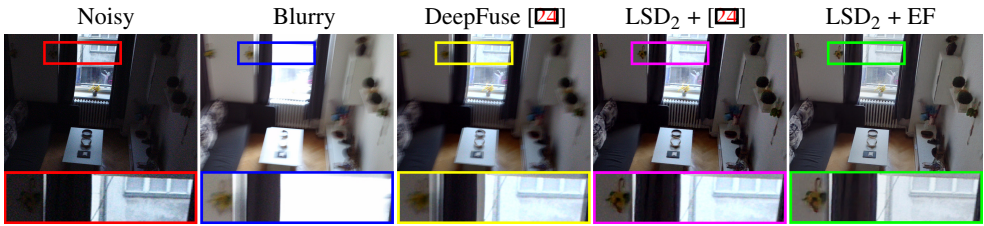

\subsection{Exposure Fusion}
LSD$_2$ network produces a sharp version of the long exposure image that is aligned with the short exposure image. Thus, the short exposure image and the LSD$_2$ output would be suitable inputs for an exposure fusion method, such as DeepFuse \cite{prabhakar2017deepfuse}, which assume that the input images are not blurry or misaligned. Fig.~\ref{fig:exposure_fusion} shows the result of DeepFuse when using a pair of noisy and blurry images as input. The results are significantly improved when DeepFuse is used together with LSD$_2$.


We also trained a second network for exposure fusion since DeepFuse \cite{prabhakar2017deepfuse} does not consider the noise and color distortions of the short exposure image. Details of the network architecture and training are given in the supplementary material. The training was done using similar synthetic long and short exposure image pairs as described in Sections \ref{sec:saturation} and \ref{sec:color_distortion}. This time the random number $s$ was uniformly sampled from the interval $[1/3,3]$ and the ground truth target is the original image, which has not been scaled by $s$ and is presumably taken with "good exposure". This type of approach differs from existing methods, which often use hand-crafted features and assume that ground truth targets are not available. 

Figs \ref{fig:teaser} and \ref{fig:exposure_fusion} show that we get higher dynamic range and better reproduction of colors and brightness than in either one of the single-exposure input images. Notice also the lack of details in the dark areas of the DeepFuse output (see e.g. the curtains). The main purpose of this experiment is to demonstrate the suitability of LSD$_2$ approach for handheld high-dynamic range imaging with smartphones. A more comprehensive evaluation of different exposure fusion techniques is left for future work.

\section{Conclusion}
We proposed a CNN-based joint image denoising and deblurring method called LSD$_2$. It recovers a sharp and noise-free image given a pair of short and long exposure images. Its performance exceeds the conventional single-image denoising and deblurring methods on both static and dynamic scenes. Furthermore, LSD$_2$ compares favorably with existing multi-image approaches. Unlike previous methods that utilize pairs of noisy and blurry images, LSD$_2$ does not rely on any existing denoising algorithm. Moreover, it does not expect the input images to be pre-aligned. Finally, we demonstrated that the LSD$_2$ output makes exposure fusion possible even in the presence of motion blur and misalignment.


\section*{Acknowledgement}
The authors would like to thank Business Finland for the financial support of this research project (grant no. 1848/31/2015). J. Matas was supported by  project CZ.02.1.01/0.0/0.0/1601\\9/0000765 Research Center for Informatics.

\bibliography{main}

\newpage
\section*{ }
\section*{Supplementary material}

\begin{figure}[!hb]
  \centering
  \begin{overpic}[width=1.0\linewidth]{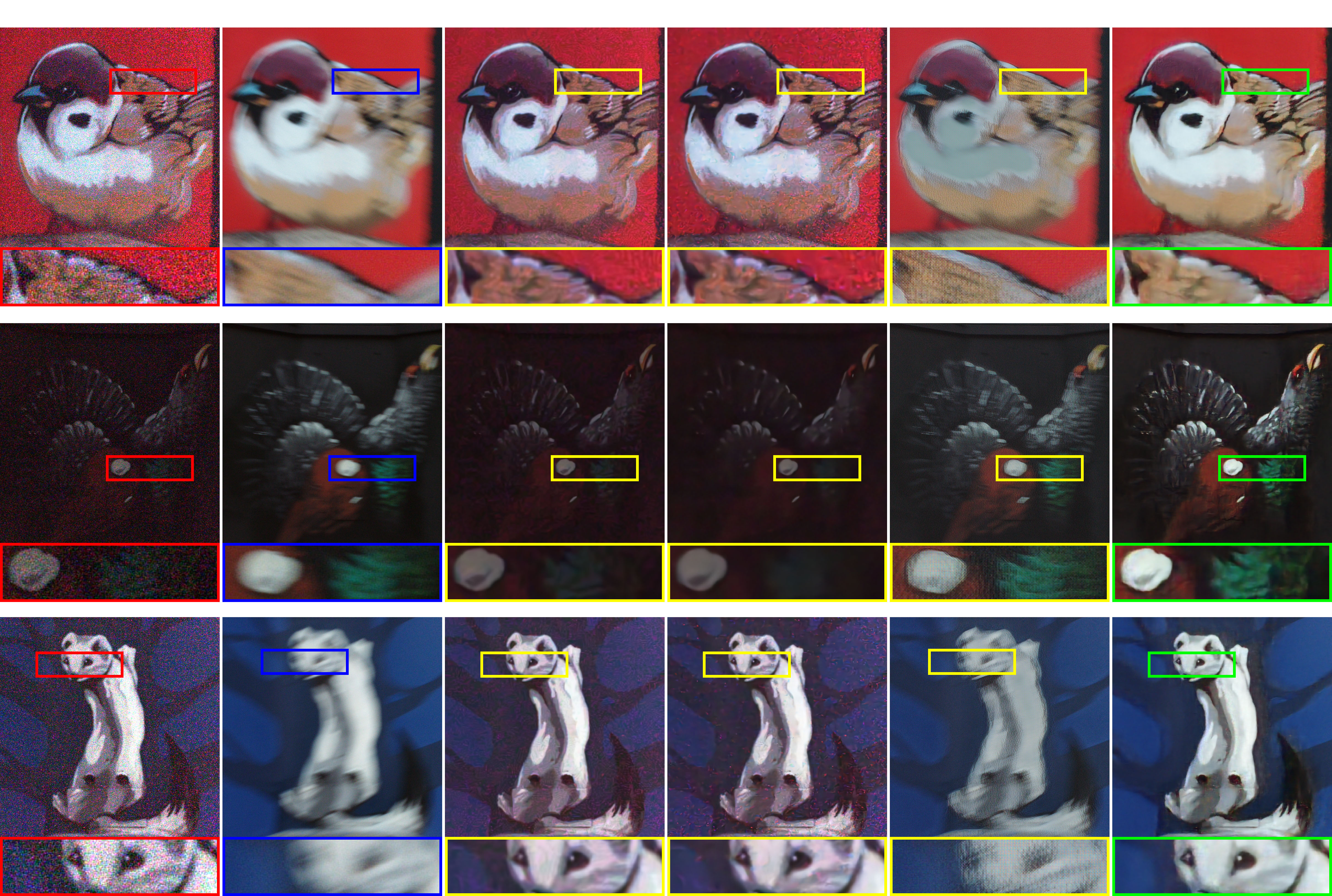}
  \put (6.0,66.0) {\small Noisy}
  \put (22.0,66.0) {\small Blurry}
  \put (36.5,66.0) {\small BM3D \cite{dabov2007image}}
  \put (51.0,66.0) {\small FDnCNN \cite{zhang2017beyond}}
  \put (67.0,66.0) {\small DeblurGAN \cite{DeblurGAN}}
  \put (89.3,66.0) {\small LSD$_2$}
  \end{overpic}
  \caption{Static scene performance (\textit{sparrow, capercaillie, weasel}).}
  \label{fig:real2}
\end{figure}

\begin{figure}
  \centering
  \begin{overpic}[width=1.0\linewidth]{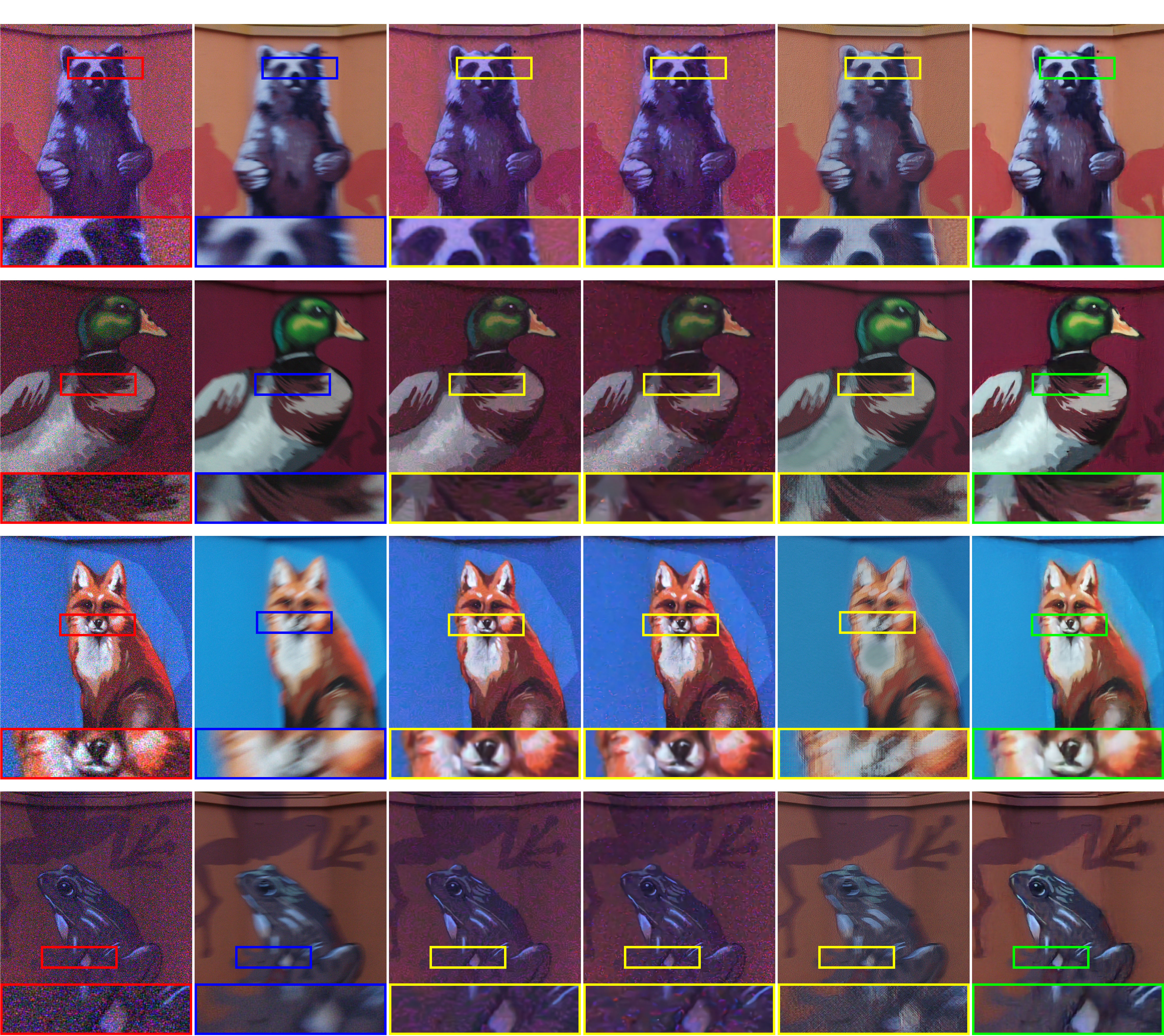}
  \put (6.0,87.9) {\small Noisy}
  \put (22.0,87.9) {\small Blurry}
  \put (36.5,87.9) {\small BM3D \cite{dabov2007image}}
  \put (51,87.9) {\small FDnCNN \cite{zhang2017beyond}}
  \put (67.0,87.9) {\small DeblurGAN \cite{DeblurGAN}}
  \put (89.3,87.9) {\small LSD$_2$}
  \end{overpic}
  \caption{Static scene performance (\textit{bear, duck, fox, frog}).}
  \label{fig:real1}
\end{figure}

\begin{figure}
  \centering
  \begin{overpic}[width=1.0\linewidth]{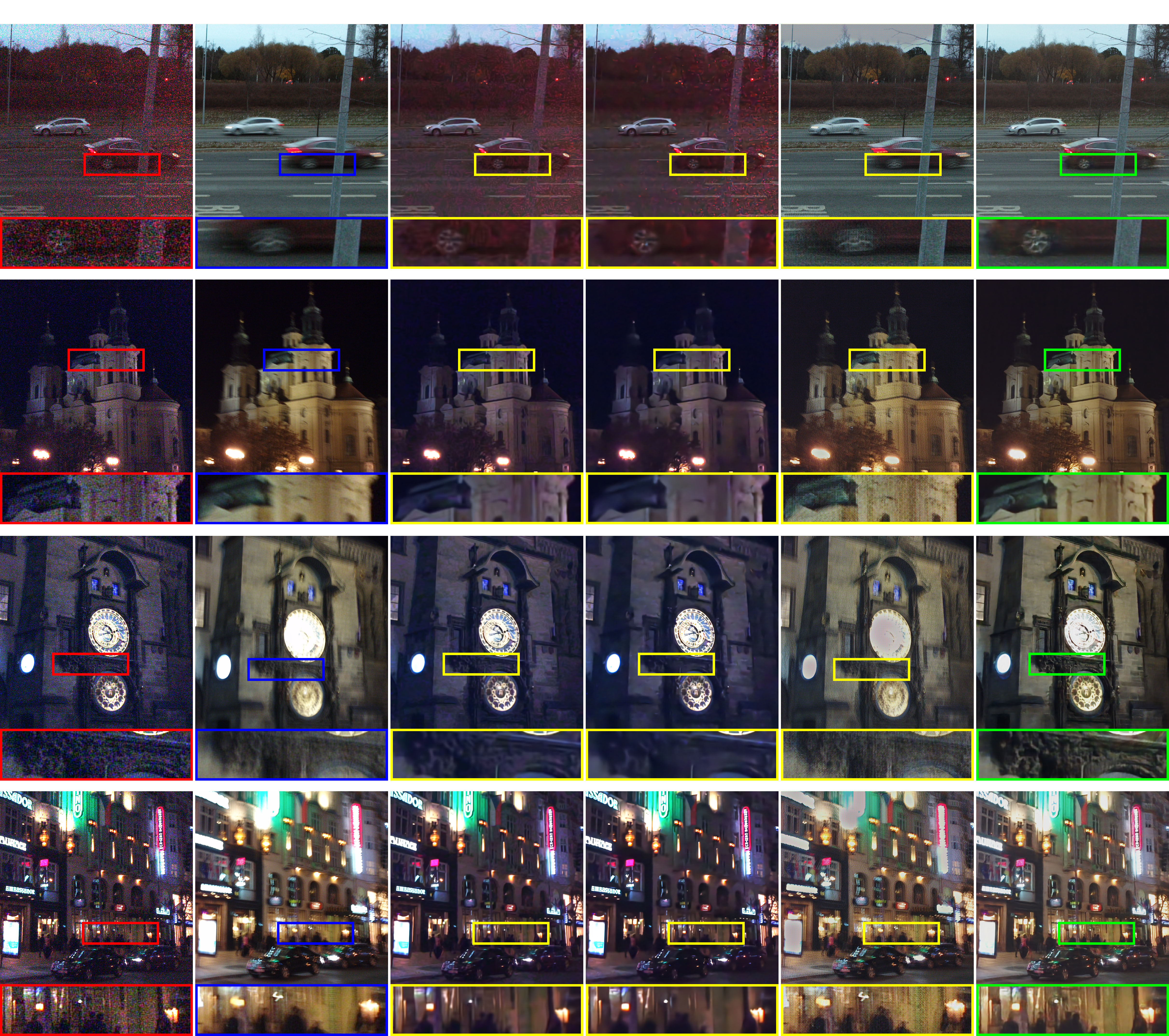}
  \put (6.0,87.6) {\small Noisy}
  \put (22.0,87.6) {\small Blurry}
  \put (36.5,87.6) {\small BM3D \cite{dabov2007image}}
  \put (51,87.6) {\small FDnCNN \cite{zhang2017beyond}}
  \put (67.8,87.6) {\small DeblurGAN \cite{DeblurGAN}}
  \put (89.3,87.6) {\small LSD$_2$}
  \end{overpic}
  \caption{Dynamic scene performance and low-light performance including saturated pixels (\textit{cars, church, clock, street}).}
  \label{fig:real3}
\end{figure}

\begin{figure}
  \centering
  \begin{overpic}[width=1.0\linewidth]{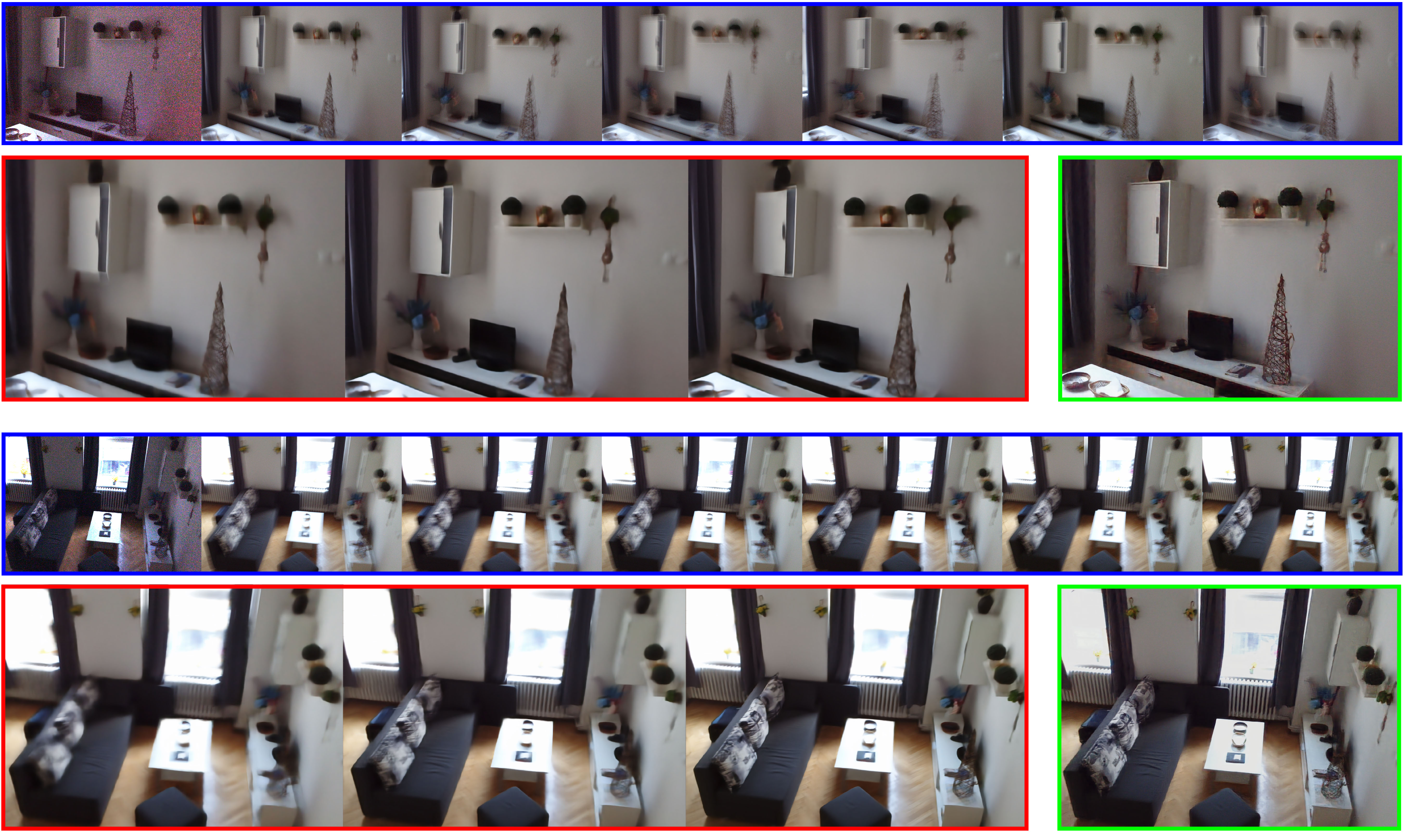}
  \end{overpic}
  \caption{A comparison with Aittala and Durand \cite{aittala2018burst}. A noisy image and a burst of blurry images (blue). The results of \cite{aittala2018burst} obtained using a growing number of input images: 1, 3 and 6 (red). The result of LSD$_2$ (green). Notice that LSD$_2$ is able to recover more details (see for example the cone shaped wooden ornament in the first test case).}
  \label{fig:aittala}
\end{figure}

\begin{figure}
  \centering
  \begin{overpic}[width=1.0\linewidth]{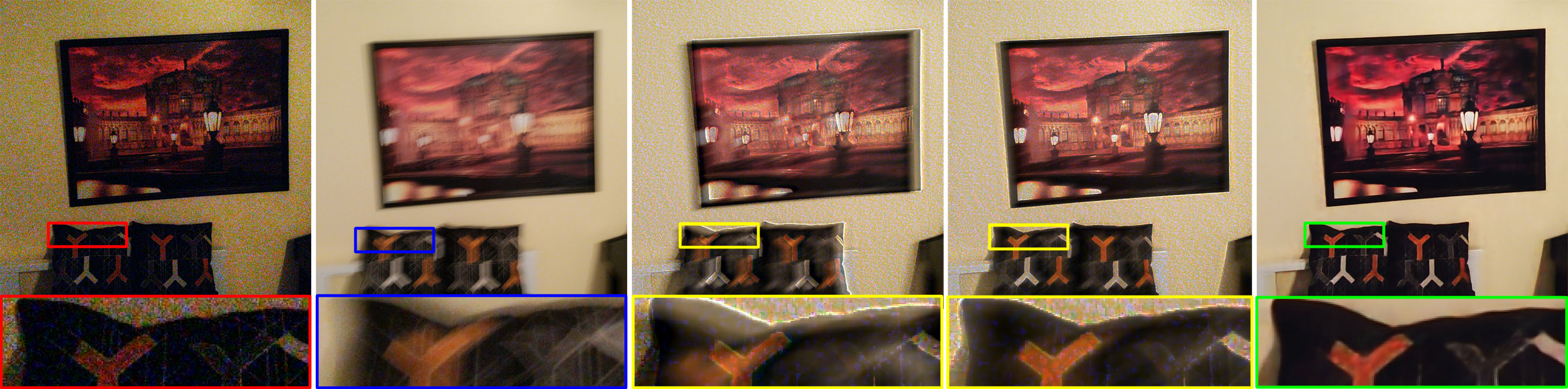}
  \put (7.0,25.8) {\small Noisy}
  \put (27.0,25.8) {\small Blurry}
  \put (43.2,28.2) {\small Yuan \etal \cite{yuan2007image}}
  \put (44.0,25.8) {\small (misaligned)}
  \put (62.5,28.2) {\small Yuan \etal \cite{yuan2007image}}
  \put (61.5,25.8) {\small (manually aligned)}
  \put (87.3,25.8) {\small LSD$_2$}
  \end{overpic}
  \caption{A comparison of LSD$_2$ and Yuan \etal \cite{yuan2007image}.}
  \label{fig:yuan}
\end{figure}

\begin{figure}
  \centering
  \begin{overpic}[width=1.0\linewidth]{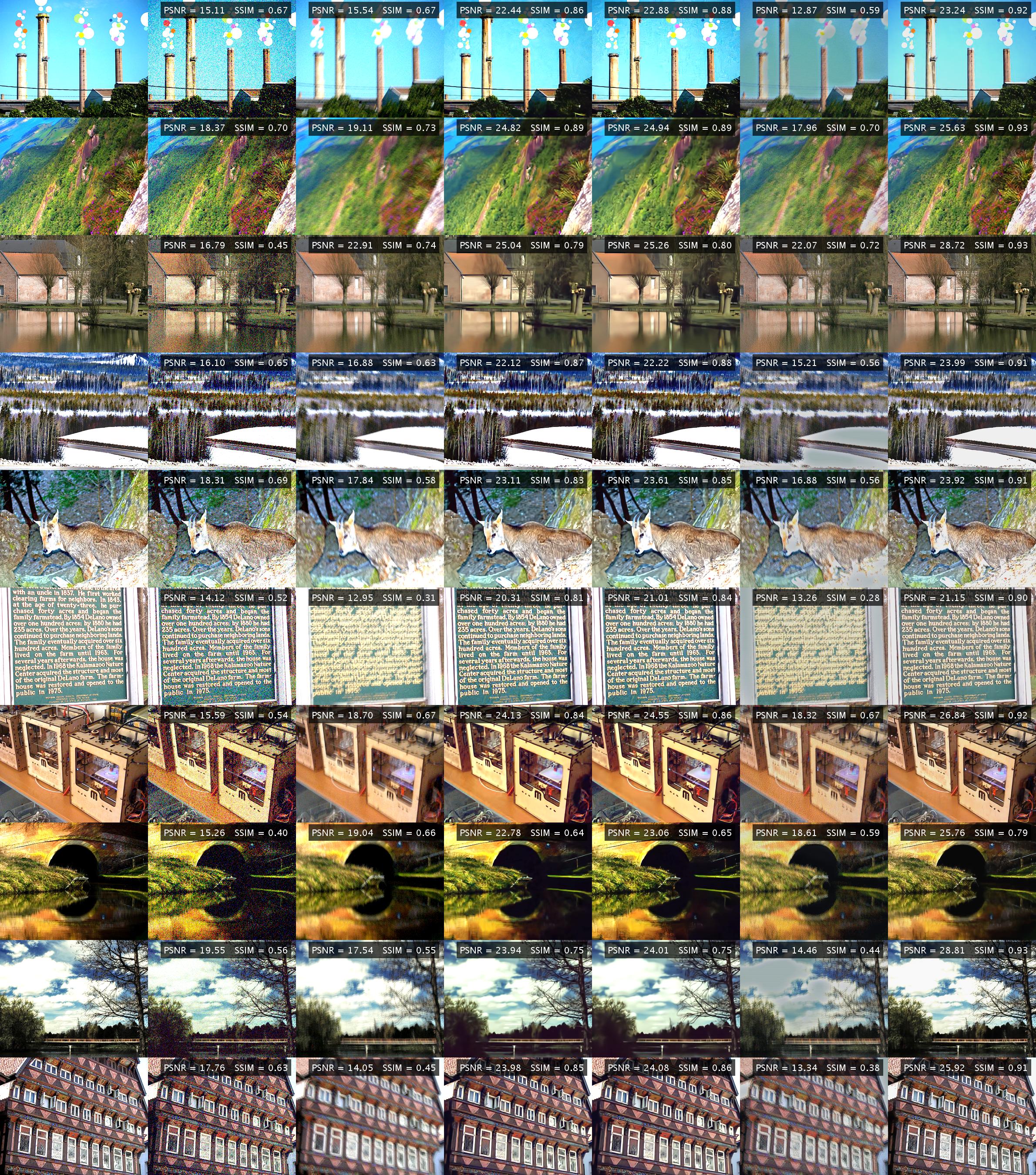}
  \put (4.0,100.7) {\small Sharp}
  \put (16.7,100.7) {\small Noisy}
  \put (28.7,100.7) {\small Blurred}
  \put (42.9,100.7) {\small \cite{dabov2007image}}
  \put (55.1,100.7) {\small \cite{zhang2017beyond}}
  \put (67.6,100.7) {\small \cite{DeblurGAN}}
  \put (80.0,100.7) {\small LSD$_2$}
  \end{overpic}
  \caption{Results on synthetically corrupted images (1-10). Noisy images and the results of \cite{dabov2007image} and \cite{zhang2017beyond} have been normalized so that the mean intensity of each color channel matches the blurred image.}
  \label{fig:synthetic1}
\end{figure}

\begin{figure}
  \centering
  \begin{overpic}[width=1.0\linewidth]{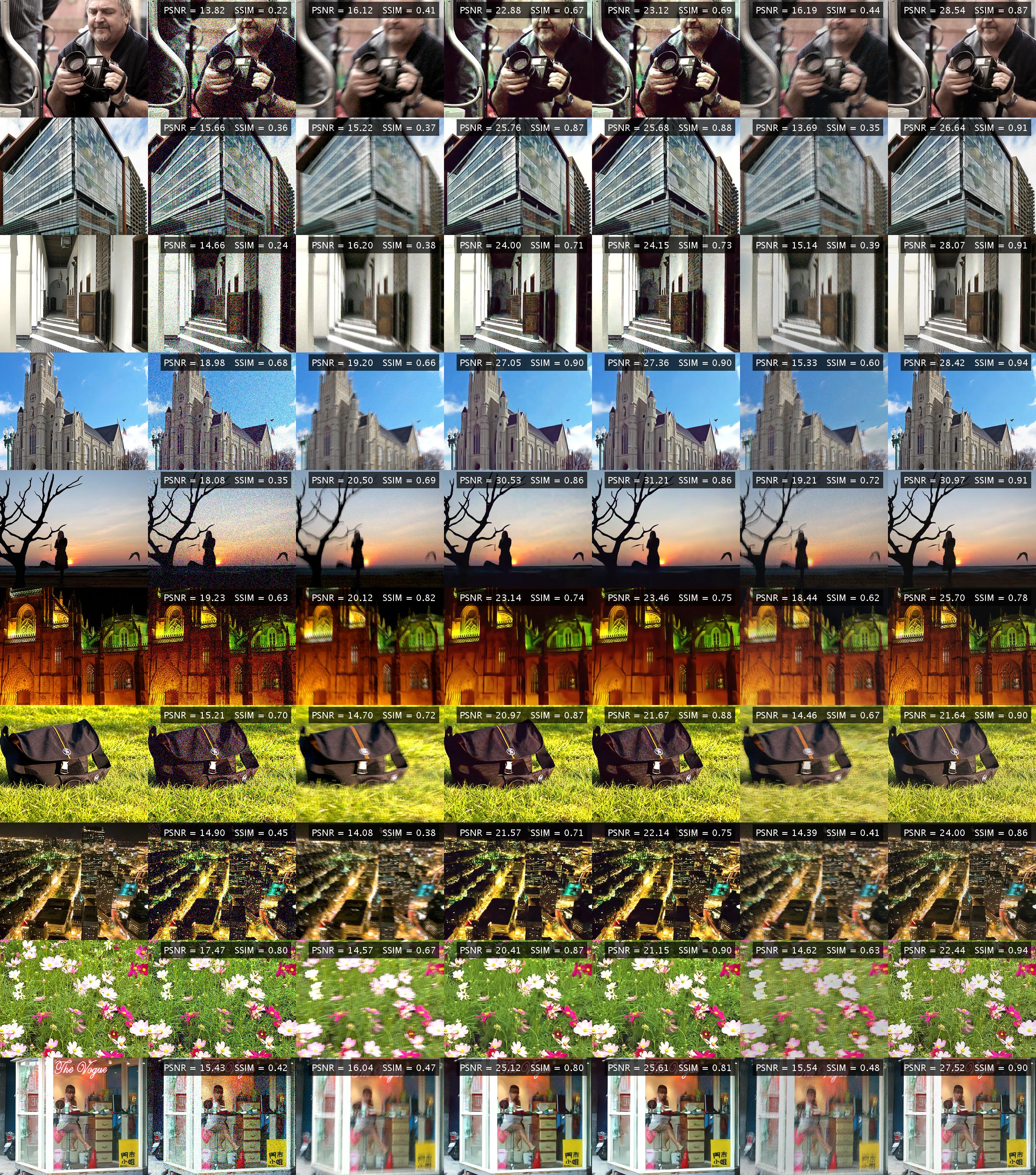}
  \put (4.0,100.7) {\small Sharp}
  \put (16.7,100.7) {\small Noisy}
  \put (28.7,100.7) {\small Blurred}
  \put (42.9,100.7) {\small \cite{dabov2007image}}
  \put (55.1,100.7) {\small \cite{zhang2017beyond}}
  \put (67.6,100.7) {\small \cite{DeblurGAN}}
  \put (80.0,100.7) {\small LSD$_2$}
  \end{overpic}
  \caption{Results on synthetically corrupted images (11-20). Noisy images and the results of \cite{dabov2007image} and \cite{zhang2017beyond} have been normalized so that the mean intensity of each color channel matches the blurred image.}
  \label{fig:synthetic2}
\end{figure}

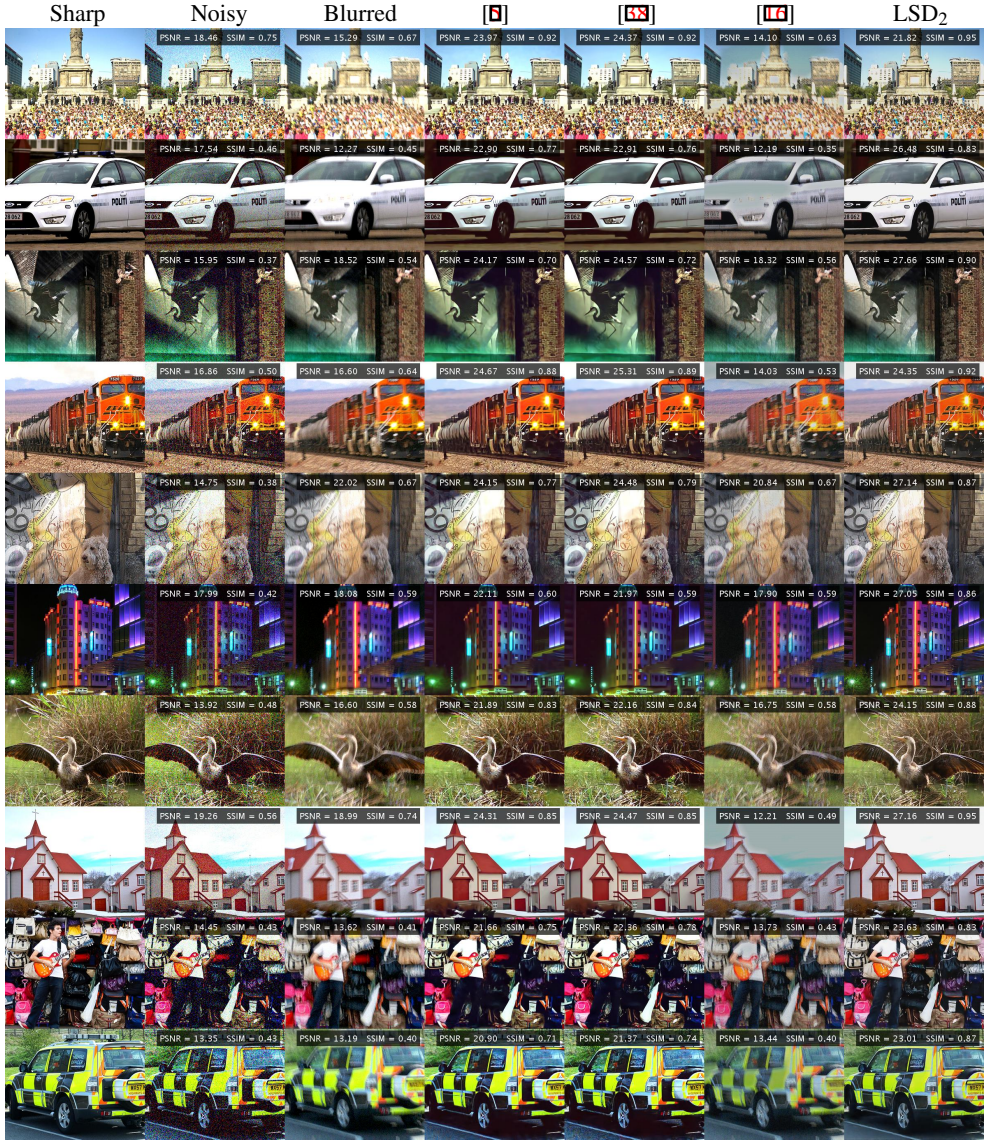
\begin{figure}
  \centering
  \begin{overpic}[width=1.0\linewidth]{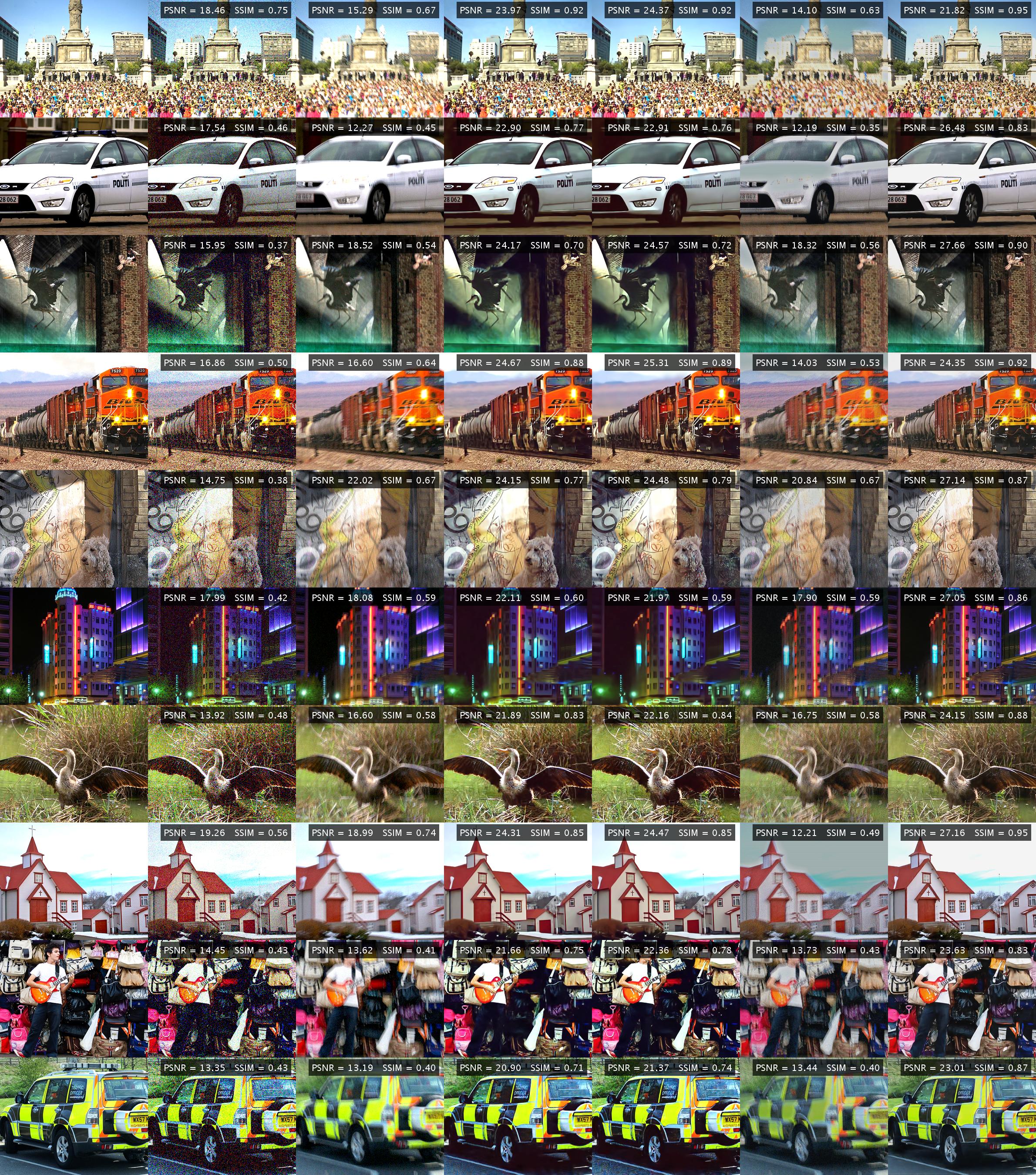}
  \put (4.0,100.7) {\small Sharp}
  \put (16.7,100.7) {\small Noisy}
  \put (28.7,100.7) {\small Blurred}
  \put (42.9,100.7) {\small \cite{dabov2007image}}
  \put (55.1,100.7) {\small \cite{zhang2017beyond}}
  \put (67.6,100.7) {\small \cite{DeblurGAN}}
  \put (80.0,100.7) {\small LSD$_2$}
  \end{overpic}
  \caption{Results on synthetically corrupted images (21-30). Noisy images and the results of \cite{dabov2007image} and \cite{zhang2017beyond} have been normalized so that the mean intensity of each color channel matches the blurred image.}
  \label{fig:synthetic3}
\end{figure}

This document contains additional examples from the same datasets shown in the paper. Images are best viewed electronically and zoomed-in. Figures \ref{fig:real2} - \ref{fig:yuan} show the results on realworld images (real motion blur and noise). Figures \ref{fig:synthetic1} - \ref{fig:synthetic3} show the results on synthetically corrupted images. Additional details of the exposure fusion method are given at the end of this document.

\vspace{8mm}


\clearpage

\section*{Exposure Fusion}
The proposed exposure fusion network takes a pair of short and long exposure images as input. Let $I_S$ and $I_L$ denote the short and long exposure images, respectively. In our case, $I_L$ is produced by the LSD$_2$ method. The output of the exposure fusion network is a weight map $W$, which is used to produce the fused image

\begin{equation}
    \hat{I}_F(i,j,k) = W(i,j) \cdot I_L(i,j,k) + [1-W(i,j)] \cdot I_S(i,j,k),
\end{equation}

where $(i,j,k)$ refers to pixel $(i,j)$ in the $k$-th color channel. We then compute the mean squared error loss given the ground truth image $I_F$, presumably taken with "good exposure". In the following sections, we provide details of the network architecture and training.

\subsection*{Architecture}
The network consists of 7 convolutional layers connected in a sequential manner. The input of the network is a pair of short and long exposure images $I_S$ and $I_L$ (stacked). The output is a weight map $W$ with the same size as the input images (single channel). All convolutional layers use a 3 $\times$ 3 window, except the last layer, which is a 1 $\times$ 1 convolution. The number of feature maps is 16 for the layers 1, 2, 5 and 6, and 32 for the layers 3 and 4. Even though the network is very simple, it produces surprisingly good results as shown in Fig. 7 of the main paper. We note that alternative network architectures might provide further improvements.

\subsection*{Training}
The network was trained on 50k images taken from an online image collection \cite{huiskes2010new}. The training was done using synthetic long and short exposure image pairs as described in Sections 3.1.1 and 3.1.2 of the main paper. The resolution of the images was 270 $\times$ 480 pixels. We used the Adam \cite{kingma2015j} optimizer. The learning rate was set to 0.00002 and the network was trained for 5 epochs.

\end{document}